# Facial beauty prediction fusing transfer learning and broad learning system

Junying Gan[1] · Xiaoshan Xie[1] · Yikui Zhai[1] · Guohui He[1] · Chaoyun Mai[1] · Heng Luo[1]

**Abstract**
Facial beauty prediction (FBP) is an important and challenging problem in the fields of computer vision and machine learning. Not only it is easily prone to overfitting due to the lack of large-scale and effective data, but also difficult to quickly build robust and effective facial beauty evaluation models because of the variability of facial appearance and the complexity of human perception. Transfer Learning can be able to reduce the dependence on large amounts of data as well as avoid overfitting problems. Broad learning system (BLS) can be capable of quickly completing models building and training. For this purpose, Transfer Learning was fused with BLS for FBP in this paper. Firstly, a feature extractor is constructed by way of CNNs models based on transfer learning for facial feature extraction, in which EfficientNets are used in this paper, and the fused features of facial beauty extracted are transferred to BLS for FBP, called E-BLS. Secondly, on the basis of E-BLS, a connection layer is designed to connect the feature extractor and BLS, called ER-BLS. Finally, experimental results show that, compared with the previous BLS and CNNs methods existed, the accuracy of FBP was improved by E-BLS and ER-BLS, demonstrating the effectiveness and superiority of the method presented, which can also be widely used in pattern recognition, object detection and image classification.

## 1 Introduction

Since Plato proposed the concept of aesthetics, research has been conducted in the fields of philosophy, psychology, and medicine to explore the nature of beauty and the criteria for evaluating it, but there is still no scientific definition. Facial beauty prediction (FBP) is a frontier topic in artificial intelligence of the nature and laws of human cognition, which is the study of how to make computers have the ability to judge or predict the beauty attractiveness of human faces similar to humans. But there is no clear and universal definition of facial beauty, which makes automatic FBP more challenging. Therefore, research of FBP is scientifically important for understanding the perception mechanism of human brain and simulating human intelligence. Simultaneously, exploring how to better interpret, quantify and predict beauty will help people understand and describe beauty more scientifically and objectively, further promoting the rapid development of related industries, such as makeup evaluation (Wei et al. 2022), makeup transfer (Wan et al. 2022), personalization recommendation (Lin et al. 2019a, b) and cosmetic surgery planning (Xie et al. 2015). In recent years, scholars have been working hard to explore deep learning and use it for FBP. Liu et al. (2019) proposed a method for understanding facial beauty via deep facial features, in which facial features are extracted by LightCNN and facial beauty is evaluated by random forest, revealing the importance of deep features in understanding facial beauty, and laying the foundation for quantitative analysis of FBP. At the same time, adaptive attribute-aware convolutional neural network of FBP is proposed (Lin et al. 2019a, b), which adaptively uses attribute-aware as additional input of model by modulating the filters of the network. And a pseudo attribute-aware convolutional neural network was proposed also (Lin et al. 2019a, b), which learned the input pseudo attribute-aware by a lightweight pseudo attribute distiller, and effectively improves the performance of FBP. However, FBP still suffers from insufficiently supervised information and is prone to overfitting because of the lack of large-scale valid data, so that it is difficult to construct an effective and robust beauty assessment model.

The training of convolutional neural networks often requires a large amount of data as training samples, while in reality a large number of labeled training samples are lacking and difficult to obtain. How to train a robust network when the training sample is insufficient need be solved. Transfer learning is a good option. In recent years, transfer learning has attracted extensive attention and research from industry and scholars. Some work (Agarwal et al. 2021; Zhuang et al.2019) reviewed transfer learning, explaining the definition and classification of transfer learning in detail. Transfer learning starts by unfreezing the fully connected layers of CNN and uses the frozen part of CNN as a fixed feature extractor for new dataset. At the same time, the model trained on the large dataset is frosted by thawing the shallow layer of CNN, and the deep part of CNN is trained with new dataset to improve the performance of the model and prevent overfitting. Xu et al. (2018) first proposed transferring the deep rich features in the pre-trained model to the Bayesian ridge regression algorithm for FBP. Our group utilized multiscale CNN, transfer learning and max-feature-map as activation function to solve FBP problem, through integrating different scales features to get good results (Zhai et al. 2019). Although these methods have achieved better results, they rely heavily on high-performance hardware devices and spend a lot of training time. In addition, in order to improve the generalization ability and accuracy of models, it needs to be retrained on the newly added data so that a lot of computer resources and time are wasted.

The training of convolutional neural networks relies on high-performance equipment and takes a lot of time. Broad learning system (BLS) was proposed to address these problems (Chen and Liu 2018), which is an efficient incremental learning system without deep architecture. In recent years, the emergence of BLS is moving towards establishing more efficient and effective machine learning methods (Gong et al. 2021). Zhang et al. (2019) proposed a face recognition method based on BLS with feature block, demonstrating how face recognition with the help of BLS is not affected by the number of facial features in strong illumination and occlusion, and hold high accuracy. And a new FBP approach was designed by our group based on local feature fusion and BLS, in which the fused features of facial beauty are extracted by 2D dimensional principal component analysis, and these features are input into BLS for FBP, greatly reducing training time (Zhai et al. 2020). Furthermore, a new method was designed for facial expression recognition in human robot interaction based on enhanced broad Siamese network (Li et al. 2021), which efficiently decreased consumption of computing time and memory resources. But the accuracy of FBP is much lower than that of methods based on deep convolutional neural networks because of its insufficient feature extraction ability.

To solve the problems above, we propose a new idea to integrate transfer learning and BLS in this paper, which can improve the training speed of the model and ensure the accuracy of FBP. Firstly, EfficientNets (Tan and Le 2019) are used as the backbone network, and all the convolutional layers are congealed, the weights are transferred from ImageNet-1 k, which are applied as feature extractor to extract facial features that would be transferred to BLS for FBP, called E-BLS. Secondly, based on E-BLS, a connection layer is designed to connect the feature extractor and BLS, called ER-BLS. In the connection layer, facial features were performed by global average pooling, batch normalization and regularization operations, and were activated by radial basis function (RBF).

We implemented extensive experiments on SCUT-FBP5500 (Liang et al. 2018) database and the Large Scale Asian Female Beauty Dataset (LSAFBD) (Zhai et al. 2016) to study the properties of E-BLS and ER-BLS. Experimental results show that E-BLS and ER-BLS presented achieve better results and outperform previous BLS methods and CNNs. Meanwhile, our methods were compared with the state-of-the-art related methods on SCUT-FBP5500 and LSAFBD, further proofing the effectiveness and superiority of the methods proposed.

The main contributions of this work are presented as follows:

1. We present a new idea to solve the problems of overfitting and slow training speed of FBP, by integrating transfer learning and BLS.
2. We instantiate two methods to fuse transfer learning and BLS, i.e., E-BLS and ER-BLS, in which the accuracy and training speed of FBP are better balanced.
3. Compared with BLS and the other methods for FBP, extensive experimental results demonstrate the superiority and effectiveness of the methods proposed, which can also be proverbially applied in pattern recognition, object detection and image classification.

The remaining content is arranged as follows: Sect. 2 outlines the related works and Sect. 3 describes the overall schemes. Section 4 analyzes experiments and compares the performance of the proposed methods with the other existing methods. Section 5 concludes this work.

## 2 Related works

### 2.1 Transfer Learning

Currently, most of the databases for FBP are small-scale. Not only it is prone to overfitting, but also has slow training speed, when CNNs are used directly to train FBP models. Transfer learning improves the learning effect of the learners in the target domain by transferring the prior knowledge of the relevant source domain. It not only enables direct model migration by way of trained CNNs to avoid retraining large-scale deep networks, but also improves the stability and generalization ability of network models.

In order to consider the correlation between tasks, our group proposed a multi-task Transfer Learning for FBP (2 M BeautyNet), in which gender recognition was taken as an auxiliary task and FBP was taken as the main task, by way of information sharing between multiple tasks to achieve fine results (Gan et al. 2020a, b). Before long, Vahdati and Suen (2022) adopted transfer and multi-task learning for FBP, which used gender recognition and ethnicity recognition as accessorial tasks to improve the performance of FBP. Bougourzi et al. (2022) proposed a two-branch architecture (REX-INCEP) based on ResneXt and Inception, by migrating trained weights on large-scale datasets, combining robust losses and ensemble regression to achieve better results on SCUT-FBP5500. Meanwhile, Dornaika and Moujahid (2022) presented Multi-Similarity Metric Fusion Manifold Embedding (MSMFME) for FBP, by migrating the weights trained on a large amount of unlabeled face data. Therefore, the problem of insufficient supervision information for FBP can be solved.

We build our feature extractor via EfficientNets (Tan and Le 2019) and transfer its weights trained on ImageNet-1k. EfficientNets are a family of models that optimize floating point operations and parameter efficiency. They optimize computational complexity and parameters by balancing the scaling multipliers $(d, r, w)$ on three dimensions: depth, width, and resolution. The scaling criterion is.

$$\begin{aligned} &\text{depth}: d = \alpha^\lambda \\ &\text{width}: w = \beta^\lambda \\ &\text{resolution}: r = \gamma^\lambda \\ &\text{s.t.}\ \alpha \cdot \beta^2 \cdot \gamma^2 \approx 2; \\ &\alpha \geq 1, \beta \geq 1, \gamma \geq 1 \end{aligned} \quad (1)$$

where $\alpha$, $\beta$, $\gamma$ are constants determined by a small grid search, which specify how to assign these extra resources of network width, depth, and resolution, respectively. Intuitively, $\lambda$ is a user-specified coefficient controlling how many resources are available for model scaling, and the speed of model operation is proportional to $d$, $w^2$, $r^2$. Compared with the other deep convolutional neural networks, EfficientNets have better tradeoff in terms of computing speed and accuracy.

### 2.2 Broad learning system

Training of CNNs requires a lot of time and high-performance equipment, and a number of works have shown that this problem can be solved by way of BLS, a high-speed learning system without deep architecture (Chen et al. 2019; Zhang et al. 2019; Zhai et al. 2020; Li et al. 2021; Chang and Chun 2022). Zhai et al. (2020) designed a new FBP architecture via BLS, by combining local feature fusion and 2D principal component analysis, training time was effectively reduced while maintaining good accuracy. Ranjana et al. (2022) proposed Broad Learning and Hybrid Transfer Learning System for face mask detection, in which good results were obtained.

BLS contains three essential parts: mapping feature nodes, enhancement feature nodes and output layer. Above all, images are mapped to feature nodes with random weights. Secondly, feature nodes are mapped to enhancement feature nodes with random weights. Finally, outputs of BLS are computed through mapping feature nodes and enhancement feature nodes. The detailed process of BLS algorithm is as follows.

Firstly, the $i$th ($i = 1,..., n$) group of feature nodes generated by the feature mapping $\phi_i$ is obtained by

$$Z_i = \phi_i(XW_{ei} + b_{ei}), i = 1, 2, \ldots, n \quad (2)$$

where the weights $W_{ei}$ and $b_{ei}$ are random weights and bias, respectively. Whole feature nodes are denoted as $Z^n \triangleq [Z_1, Z_2, \ldots, Z_n]$.

Secondly, the $k$th ($k = 1, ..., m$) group of enhancement nodes is generated by the nonlinear activation function $\xi_k$, that is

$$H_k = \xi_k(Z^n W_{hk} + b_{hk}), k = 1, 2, \ldots, m \quad (3)$$

analogously, $W_{hk}$ and $b_{hk}$ are random samples of some certain distributions. Whole enhancement nodes are indicated as $H^m \triangleq [H_1, H_2, \cdots, H_m]$.

Finally, the output layer of BLS constructs a desired result Y. The consequent equation is

$$Y \triangleq [Z^n, H^m] W^o \tag{4}$$

where $W^o$ is the weights of the output layer. It should be noted that $W^o = [Z^n, H^m]^+ Y$ is obtained by the pseudo inverse of matrix $[Z^n, H^m]$.

An important advantage of BLS is that additional feature nodes and enhancement nodes can be dynamically added to the system, and retraining the whole system can be avoided. In incremental BLS, the $(n + 1)$-th feature mapping group nodes are added and expressed as

$$Z_{n+1} = \phi_{n+1}(XW_{e_{n+1}} + b_{e_{n+1}}) \tag{5}$$

where $W_{e_{n+1}}$ and $b_{e_{n+1}}$ are samples of some given distributions.

The corresponding $(m + 1)$-th enhancement nodes are represented as

$$H_{m+1} = \xi_{m+1}(Z^{n+1} W_{hm+1} + b_{hm+1}) \tag{6}$$

similarly, $W_{hm+1}$ and $b_{hm+1}$ are random weights and bias, respectively.

Suppose $A_n^m \triangleq [Z^n, H^m]$, the updated combined matrix and its pseudo-inverse matrix are given by

$$A_{n+1}^{m+1} = [A_n^m, Z_{n+1}, H_{m+1}] \tag{7}$$

To update the output weights $W_{n+1}^{m+1}$ from $W_n^m$, the dynamic solution could be calculated by

$$(A_{n+1}^{m+1})^+ = \begin{bmatrix} (A_n^m)^+ - DB^T \\ B^T \end{bmatrix} \tag{8}$$

$$(W_{n+1}^{m+1}) = (A_{n+1}^{m+1})^+ Y = \begin{bmatrix} (W_n^m) - DB^T Y \\ B^T Y \end{bmatrix} \tag{9}$$

where the matrices $D$ and $B$ are computed by

$$D = (A_n^m)^+ [Z_{n+1}, H_{m+1}] \tag{10}$$

$$B^T = \begin{cases} C^+, & C \neq 0 \\ (1 + D^T D)^{-1} D^T (A_n^m)^+, & C = 0 \end{cases} \tag{11}$$

$$C = [Z_{n+1}, H_{m+1}] - AD \tag{12}$$

Therefore, BLS is incrementally updated without the retraining of the whole system, greatly improving the learning efficiency of the system.

## 3 Fused strategies

### 3.1 E-BLS

The architecture of E-BLS is shown in Fig. 1. Among them, the network backbone contains the feature extractor and BLS. The feature extractor is used to extract facial features, which can be implemented by various existing CNN models. And we apply EfficientNets in our experiments. BLS is used for FBP model training and testing. The details of E-BLS are expressed in Algorithm 1.

---

**Algorithm 1** Facial Beauty Prediction with E-BLS

**Input:** training samples set $X$;
**Output:** output matrix set $Y$ and $W$;
Calculate $M = \text{Swish}(XW_e + b_e)$;
**For** $i = 0; i \leq n$ **do**
　Random $W_{ei}$, $b_{ei}$;
　Calculate $Z_i = \phi_i(MW_{ei} + b_{ei})$;
**End**
Set feature nodes group $Z^n \triangleq [Z_1, Z_2, \cdots, Z_n]$;
**For** $k = 0; k \leq m$ **do**
　Random $W_{hk}$, $b_{hk}$;
　Calculate $H_k = \xi_k(Z^n W_{hk} + b_{hk})$;
**End**
Set enhancement nodes group $H^m \triangleq [H_1, H_2, \cdots, H_m]$;
Calculate $Y \triangleq [Z^n, H^m] W^o$;
Calculate $W^o = [Z^n, H^m]^+ Y$;
**While** the training accuracy threshold is not satisfied **do**
　Random $W_{en+1}$, $b_{en+1}$;
　Calculate $Z_{n+1} = \phi_{n+1}(MW_{e_{n+1}} + b_{e_{n+1}})$;
　Random $W_{hm+1}$, $b_{hm+1}$;
　Calculate $H_{m+1} = \xi_{m+1}(Z^{n+1} W_{hm+1} + b_{hm+1})$;
　Set $A_{n+1}^{m+1} = [A_n^m, Z_{n+1}, H_{m+1}]$;
　Calculate $(A_{n+1}^{m+1})^+$ by Eq. (8, 10, 11, 12);
　Calculate $Y \triangleq [Z^{n+1}, H^{m+1}] W_{n+1}^{m+1}$;
　Calculate $(W_{n+1}^{m+1}) = (A_{n+1}^{m+1})^+ Y$;
　$n = n + 1$;
　$m = m + 1$;
**End**

---

Facial images are fed into a feature extractor, which is built on EfficientNets with transfer learning. the fused features of facial beauty are output from the last convolutional layer of EfficientNets by the following formula.

$$M = \text{Swish}(XW_e + b_e) \tag{13}$$

where $X$ represents the input images and Swish represents activation function. $W_e$ and $b_e$ are the weights and biases of EfficientNets, respectively.

We map $M$ to feature nodes $Z^n$ with random weights by formula (2). And then we map $Z^n$ to enhancement nodes $H^m$ with random weights by formula (3).

Finally, we construct a desired result $Y$ by formula (4) to assess facial beauty. If training accuracy threshold does not satisfy our expectation, we expand the feature nodes and enhancement nodes of BLS to improve accuracy and training speed by formulas (5, 6, 7, 8, 9, 10, 11, 12).

### 3.2 ER-BLS

The architecture of ER-BLS is shown in Fig. 2, which consists of a feature extractor, connection layer and BLS. We designed a connected layer to connect the feature extractor and BLS. The facial features are processed by the connected layer and then input to BLS for FBP. In the connected layer, global average pooling, batch normalization and regularization are performed to facial features, which are activated with RBF. At the same time, the connected layer can heighten the training speed of the model and avoid overfitting. The details of ER-BLS are expressed in Algorithm 2.

Similarly, images $X$ are fed into EfficientNets for features extraction and output facial features $M$ by formula (13). After processing through the connected layer, we output new facial features $M^*$ by formula (14), which are fed into BLS for FBP.

$$M^* = e^{\frac{[-\text{BN}(MW_r + b_r) * \text{BN}(MW_r + b_r)/2]}{\sqrt{2\pi}}} \tag{14}$$

where BN is Batch Normalization, $W_r$ and $b_r$ are the weights and biases of connected layer, respectively. We adopt RBF as activation function.

Analogously, we map $M^*$ to the feature nodes $Z^n$ with random weights by formula (2). Then we map $Z^n$ to the enhancement nodes $H^m$ with random weights by formula (3). Finally, we compute the output of BLS by formula (4) to assess facial beauty. If training accuracy threshold does not meet our expectation, we extend the feature nodes and enhancement nodes of the model to improve accuracy and training speed with formulas (5, 6, 7, 8, 9, 10, 11, 12).

---

**Algorithm 2** Facial Beauty Prediction with ER-BLS

**Input:** training samples set $X$;
**Output:** output matrix set $Y$ and $W$;
Calculate $M = \text{Swish}(XW_e + b_e)$;
Calculate $M^* = e^{\frac{[-\text{BN}(MW_r + b_r) * \text{BN}(MW_r + b_r)/2]}{\sqrt{2\pi}}}$;
**For** $i = 0; i \leq n$ **do**
    Random $W_{ei}$, $b_{ei}$;
    Calculate $Z_i = \phi_i(M^* W_{ei} + b_{ei})$;
**End**
Set feature nodes group $Z^n \triangleq [Z_1, Z_2, \cdots, Z_n]$;
**For** $k = 0; k \leq m$ **do**
    Random $W_{hk}$, $b_{hk}$;
    Calculate $H_k = \xi_k(Z^n W_{hk} + b_{hk})$;
**End**
Set enhancement nodes group $H^m \triangleq [H_1, H_2, \cdots, H_m]$;
Calculate $Y \triangleq [Z^n, H^m] W^o$;
Calculate $W^o = [Z^n, H^m]^+ Y$;
**While** the training accuracy threshold is not satisfied **do**
    Random $W_{en+1}$, $b_{en+1}$;
    Calculate $Z_{n+1} = \phi_{n+1}(M^* W_{e_{n+1}} + b_{e_{n+1}})$;
    Random $W_{hm+1}$, $b_{hm+1}$;
    Calculate $H_{m+1} = \xi_{m+1}(Z^{n+1} W_{hm+1} + b_{hm+1})$;
    Set $A_{n+1}^{m+1} = [A_n^m, Z_{n+1}, H_{m+1}]$;
    Calculate $(A_{n+1}^{m+1})^+$ by Eq. (8, 10, 11, 12);
    Calculate $Y \triangleq [Z^{n+1}, H^{m+1}] W_{n+1}^{m+1}$;
    Calculate $(W_{n+1}^{m+1}) = (A_{n+1}^{m+1})^+ Y$;
    $n = n + 1$;
    $m = m + 1$;
**End**

---

## 4 Experiments and results analysis

To study the properties of E-BLS and ER-BLS, firstly, the methods proposed in this paper are compared with the deep convolutional neural network methods based on Transfer Learning. Secondly, the effectiveness and superiority of E-BLS and ER-BLS have been demonstrated through numerous trials. Finally, our methods are compared with the related methods, respectively. All the experiments were conducted on a Python software platform with an Intel-i7 3.6 GHz CPU and 64 GB RAM desktop computer.

## 4.1 Experimental object

*SCUT-FBP5500 [10]* SCUT-FBP5500 is a facial beauty prediction database established by South China University of Technology. It contains 5,500 frontal face images at the resolution of 350 × 350 with different races, gender and age. Each image is rated by 60 volunteers and is labeled with a beauty score ranging from 1 to 5. The larger the score, the more attractive. We use mode as criterion, all the images are divided into five levels "1", "2", "3", "4" and "5", which is relevant to "extremely unattractive", "unattractive", "average", "attractive" and "extremely attractive", respectively. Among them, there were 76 images in level "1", 821 images in level "2", 3278 images in level "3", 1226 images in level "4" and 99 images in level "5". Figure 3 shows the distribution some beautiful levels of SCUT-FBP5500 and Fig. 4 shows some image samples of SCUT-FBP5500.

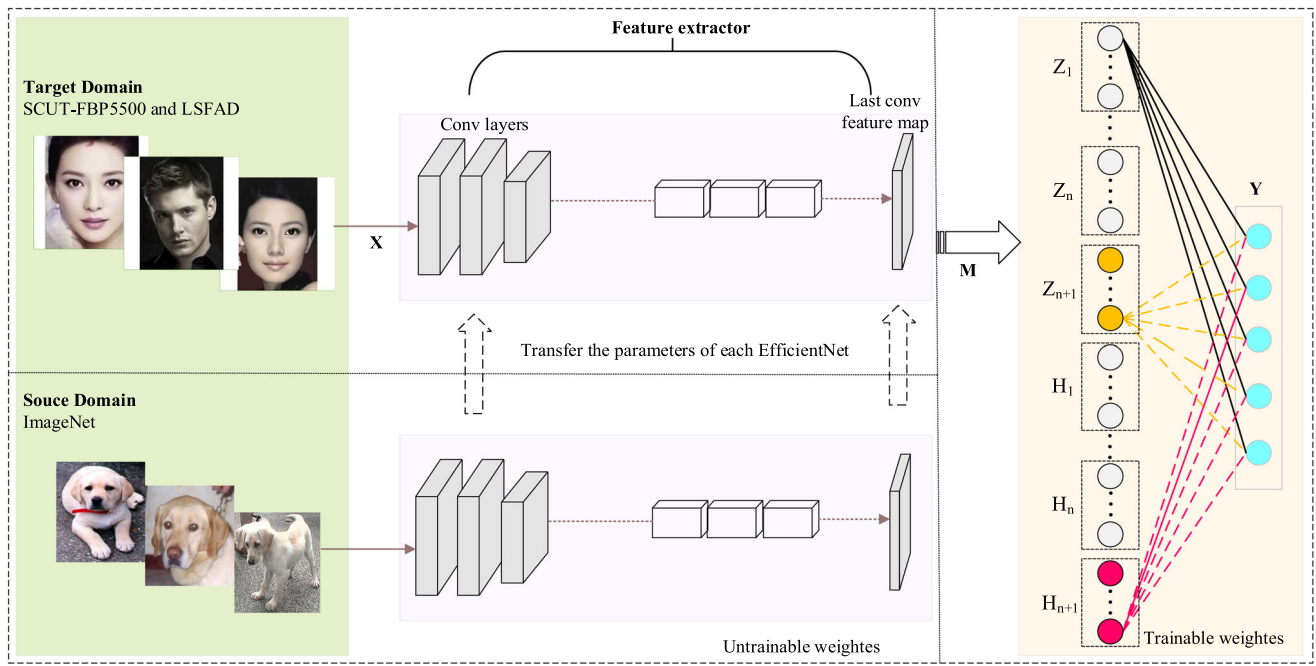

**Fig. 1** Architecture of E-BLS. It consists of a feature extractor and BLS. The feature extractor can be used to extract facial features as the input of BLS. And BLS predicts facial beauty

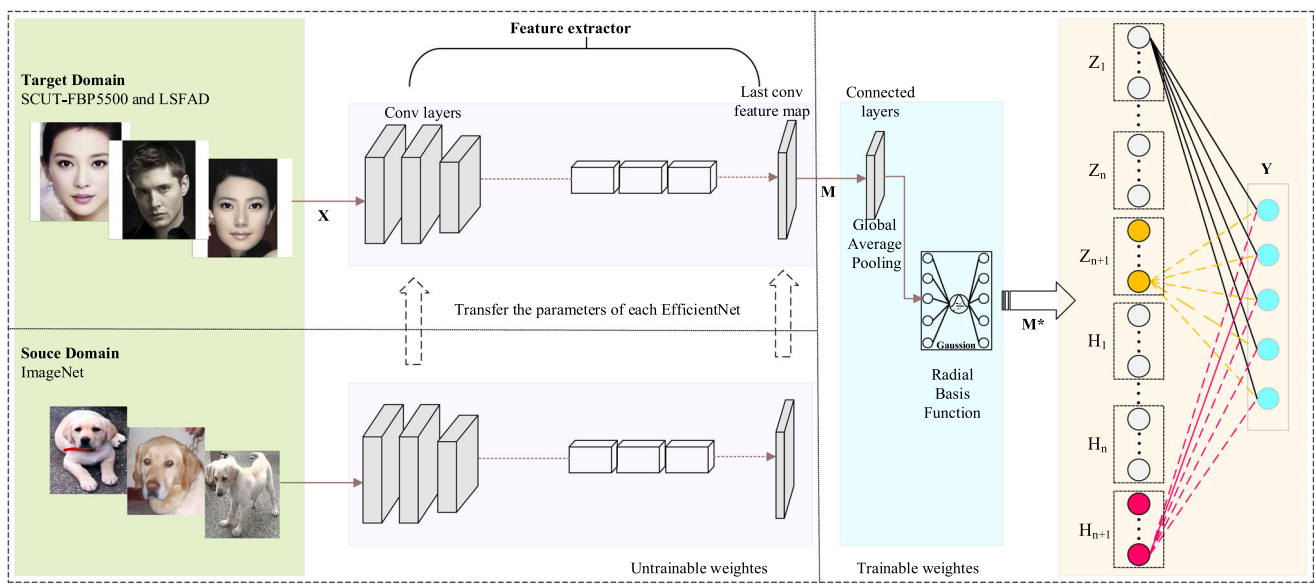

**Fig. 2** Architecture of ER-BLS. It consists of a feature extractor, connecting layer and BLS. The feature extractor can be used to extract facial features as the input of a connected layer. In the connection layer, facial features were performed by global average pooling, batch normalization and regularization operations, and were activated by RBF. BLS predicts facial beauty

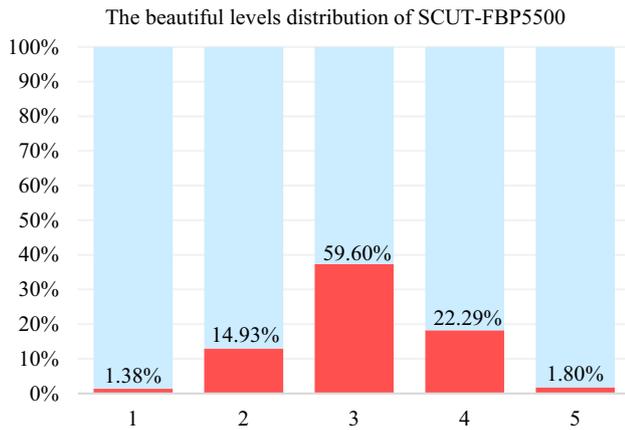

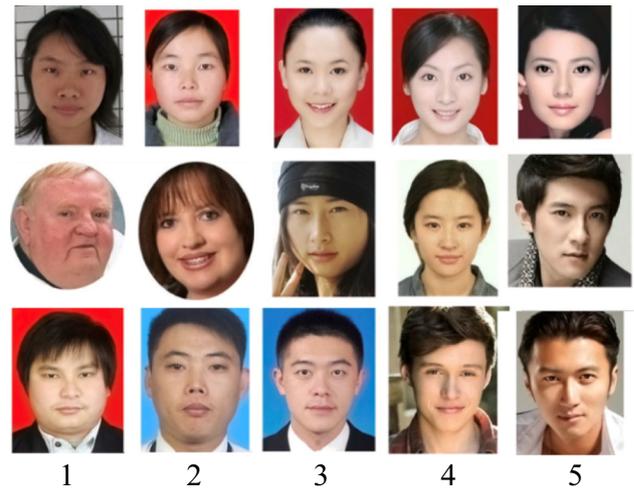

Fig. 3 The distribution of SCUT-FBP5500 with beautiful levels

Fig. 4 Facial images with different properties of SCUT-FBP5500

*LSAFBD [10]* LSAFBD is a facial beauty prediction database established by our group, which consists of 20,000 labeled images and 80,000 unlabeled images with the resolution of 144 × 144. Most facial images include variations in background, pose, and age. Each image is rated by 75 volunteers and all the images were divided into five levels, labeled as "0", "1", "2", "3" and "4", in increasing order of beauty. Among them, there were 948 images in level "0", 1148 images in level "1", 3846 images in level "2", 2718 images in level "3" and 1333 images in level "4". This paper focuses on the prediction of female beauty, and only 10,000 LSAFBD female images were used to verify the effectiveness of our methods for FBP problems. Figure 5 shows the distribution of some beautiful levels of LSAFBD and Fig. 6 shows some image samples of LSAFBD.

### 4.2 Model training and testing

These two databases are randomly divided into training set and testing set in the ratio of 8: 2 in our experiments, and experiments are conducted with cross-validation to ensure reliability. E-BLS and ER-BLS presented contain a very small number of hyperparameters. For BLS, its hyperparameters include the number of feature windows ($N1$), the number of nodes in each feature window ($N2$), and the number of enhancement nodes ($N3$). In this section, Hyperopt (Bergstra et al. 2022) was used to optimize the optimal value of $N1$, $N2$ and $N3$. To quantify the improvement effect of the presented methods, the plain BLS model was trained with the same hyperparameters. In addition, some classical deep CNNs, such as ResNet50 (He et al. 2016), InceptionV3 (Szegedy et al. 2016), DenseNet121 (Huang et al. 2017), InceptionResNetV2 (Szegedy et al. 2017), EfficientNetB7 (Tan and Le 2019), MobileNetV2 (Sandler et al. 2018), NASNet (Zoph et al. 2018), and Xception (Chollet 2017) based on Transfer Learning are also trained with the same hyperparameters. The initial learning rate is 0.001. When training accuracy does not improve for more than 3 epochs, the multiplicative factor of learning rate decay is 0.5. The batch sizes and epochs are 16 and 50, respectively. The coefficient of regularization is 0.3 and activation function utilizes linear rectification function. Meanwhile, the initial weight of these networks comes from ImageNet-1 k. Because deep CNNs are trained by Transfer Learning, all the layers except the final layer were frozen and then only the final layer was trained. When images are fed into these networks for training, the original resolution is maintained. The time, accuracy (AC) and Pearson correlation (PC) are utilized to evaluate the performance of our methods, while a short time, a high AC and PC denotes better performance. Experimental results are the average of five tests.

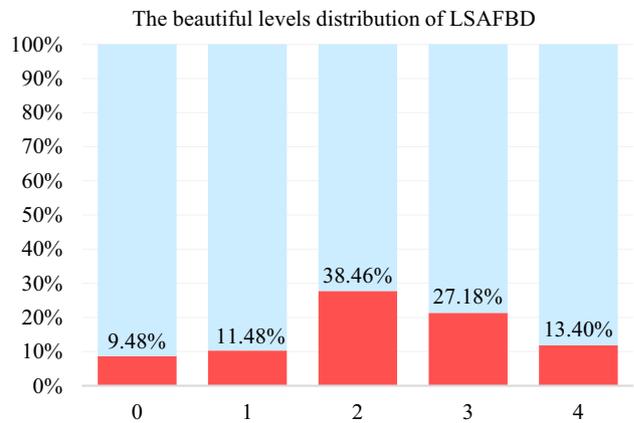

Fig. 5 The distribution of LSAFBD with beautiful levels

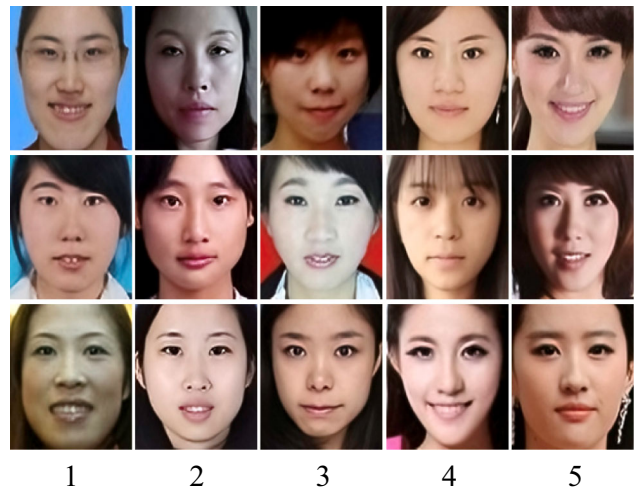

Fig. 6 Facial images with different properties of LSAFBD

Table 1 Results of FBP on SCUT-FBP5500

| Model | Training time (s) | Testing AC (%) | Training AC (%) |
|---|---|---|---|
| *E-BLS (ours)* | *1399.88* | *73.13* | *75.38* |
| BLS | 640.5 | 65.85 | 68.83 |
| NASNet | 5493.06 | 68.11 | 73.92 |
| MobileNetV2 | 4224.42 | 69.03 | 79.3 |
| DenseNet121 | 9862.52 | 70.77 | 75.97 |
| ResNet50 | 8211.02 | 71.23 | 76.63 |
| InceptionResNetV2 | 25,896.82 | 71.69 | 74.28 |
| EfficientNetB7 | 17,589.68 | 72.70 | 76.68 |
| Xception | 6770.64 | 72.98 | 75.01 |
| *ER-BLS (ours)* | *1291.83* | *74.69* | *76.76* |
| InceptionV3 | 11,630.7 | 73.16 | 78.02 |

### 4.3 Experiments on SCUT-FBP5500

In this section, we conducted extensive experiments on SCUT-FBP5500 to study the properties of E-BLS and ER-BLS. In E-BLS, Hyperopt was used to optimize the hyperparameters: $N1 = 12$, $N2 = 54$ and $N3 = 2296$ for the training of BLS and E-BLS. Experimental results are listed in Table 1. The prediction accuracy of FBP utilizing BLS directly is only 65.85%, which is the lowest among all the methods. Nevertheless, testing accuracy of E-BLS is 73.13%, which is 7.28% better than BLS, 0.43% better than EfficientNetB7, and only 0.03% lower than InceptionV3. Furthermore, training time for deep convolutional neural networks based on Transfer Learning was between 4224.42 s and 25,896.82 s, while E-BLS only needs 1399.88 s. There is no doubt that E-BLS improved the efficiency of FBP among several times and more than a dozen times while maintaining the accuracy of the model.

The training loss curve and validation loss curve of these deep CNNs are shown in Figs. 7 and 8, respectively. Each network nearly tends to converge after about 50 epochs. When we increase epochs with little performance improvement, the training time increases greatly. In addition, the loss curve of EfficientNetB7 training and validation is more stable than the other networks.

### 4.4 Experiments on LSAFBD

In this section, we continue to conduct extensive experiments to evaluate the performance and study the properties of E-BLS and ER-BLS on LSAFBD. In E-BLS and BLS, the hyperparameters are as follows: $N1$ is 25, $N2$ is 72 and $N3$ is 3088. Experimental results are listed in Table 2. Testing accuracy of FBP by BLS directly is 52.96%, which is the lowest among all the methods. Nevertheless, the testing accuracy utilizing E-BLS is 60.82%, which is 9.17% better than BLS, 2.91% better than EfficientNetB7. Its testing accuracy outperformed the convolutional neural networks. Furthermore, training time for deep convolutional neural networks based on Transfer Learning was between 7830.47 s and 36,853.53 s, while E-BLS merely needs 2300.48 s. Therefore, training efficiency of FBP with E-BLS is significantly improved.

### 4.5 Further discussion

Improved quantification of ER-BLS is listed in Tables 3 and 4, respectively. As we can see, on SCUT-FBP5500, compared with E-BLS, testing accuracy of ER-BLS is improved by 1.56% and training time is reduced by 108.05 s. Furthermore, compared with the other algorithms, accuracy of ER-BLS is improved between 1.53% and 8.84%, and the training time is shortened between 2932.59 s and 24,604.99 s.

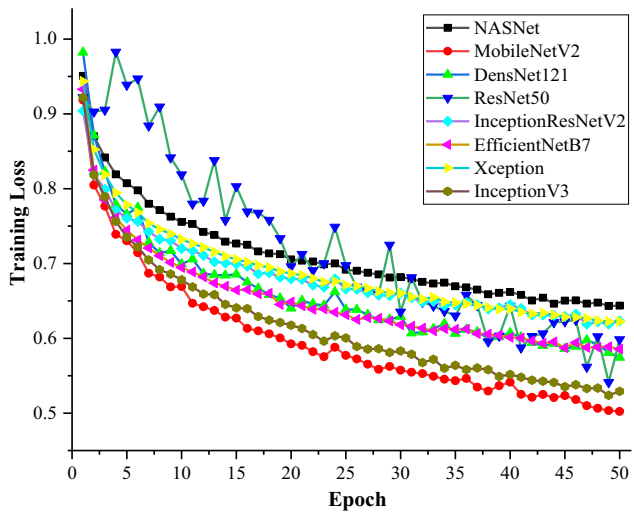

**Fig. 7** Training loss curves for each algorithm

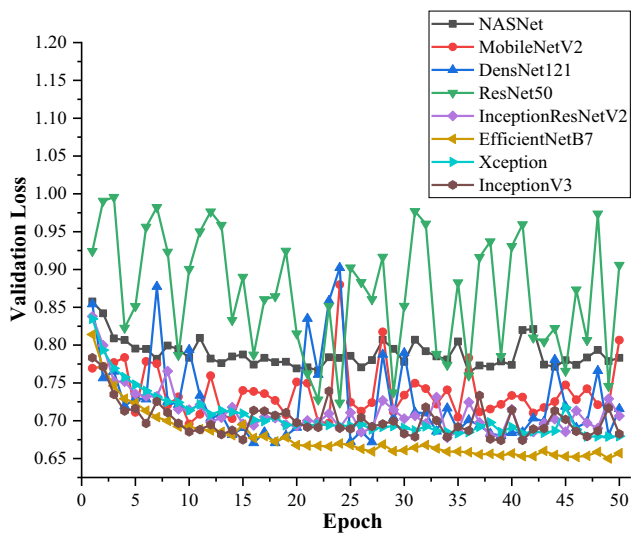

**Fig. 8** Validation loss curves of each algorithm

On LSAFBD, compared with E-BLS, testing accuracy of ER-BLS is improved by 1.31% and training time is reduced by 13.94 s. Moreover, compared with the other algorithms, accuracy of ER-BLS is improved between 2.20% and 9.17%, and training time is shortened between 5543.93 s and 34,566.99 s. The accuracy and training speed of FBP is significantly improved with ER-BLS.

ER-BLS is able to get fine accuracy with a few parameters. ER-BLS is trained with $12 \times 54$ feature nodes and 2966 enhancement nodes. These nodes directly determine the learning effect of ER-BLS. To further explore the performance of ER-BLS, studies with different mapping feature nodes and enhancement nodes were conducted. From Table 5, we can see that the more the total number of nodes, the longer the training time, but testing accuracy of ER-BLS decreases after the initial improvement. This implies that the choice of the number of nodes will become a key factor affecting the accuracy of FBP. Therefore, it is a reliable scheme to select parameters by Hyperopt.

Both the number and distribution of samples may affect the accuracy of ER-BLS. To further investigate this issue, we modified the number of training samples on LSAFBD. LSAFBD contains 10,000 images, in which 8000 images are used as training set and 2000 images as testing set. The number of images in testing set is not changed, and the number of images in training set is modified to $4000 \sim 8000$, as listed in Table 6. As can be seen from Table 6, the testing accuracy increases greatly when the number of training samples increases. Thus, we believe that the number of training samples has a large influence on the accuracy of ER-BLS.

### 4.6 Ablation study

To investigate the property of each component of our networks more clearly, EfficientNets are selected to execute our ablation studies. In the experiments, EfficientNets are set as the default backbone networks of ER-BLS. Firstly, we carry out a series of experiments on EfficientNetB7 that increases the number of training epoch and batch size gradually. Secondly, we conduct a series of experiments on ER-BLS with transfer learning or not. In the following, we conduct extensive ablation experiments to study the properties of our methods.

*Different training epoch and batch size* To explore the effect of different training epoch and batch size for FBP, we conduct experiments on EfficientNetB7 for FBP. The comparison results on SCUT-FBP5500 are shown in Figs. 9 and 10, respectively. Thus, we can draw the following conclusions:

1. 1. when training epoch is 50 and batch size is 16, the effect is the best;
2. increasing the number of training epoch and batch size will greatly increase training time, but accuracy has not been improved accordingly.

*Whether to fuse transfer learning* We explore the effect of Transfer Learning on ER-BLS. In our experiments, EfficientNets equipped with different scale are implemented as backbones. Experiments on SCUT-FBP5500 were performed in this section. As listed in Table 7, we can see that:

1. ER-BLS with transfer learning performs better than without transfer learning under the same backbone;
2. Under transfer learning, ER-BLS based on EfficientNetB7 performs better than the other backbones.

**Table 2** Results of FBP on LSAFBD

| Model | Training time (s) | Testing AC (%) | Training AC (%) |
|---|---|---|---|
| *ER-BLS (ours)* | *2286.54* | *62.13* | *72.34* |
| *E-BLS (ours)* | *2300.48* | *60.82* | *71.58* |
| BLS | 1446.37 | 52.96 | 67.89 |
| NASNet | 12,043.95 | 53.12 | 60.72 |
| MobileNetV2 | 7830.47 | 54.74 | 66.66 |
| Xception | 16,852.60 | 57.01 | 62.91 |
| InceptionResNetV2 | 36,853.53 | 57.31 | 60.91 |
| EfficientNetB7 | 21,165.61 | 57.91 | 61.32 |
| ResNet50 | 20,014.05 | 58.37 | 65.10 |
| InceptionV3 | 25,201.29 | 58.42 | 66.47 |
| DensNet121 | 19,520.28 | 59.93 | 63.54 |

**Table 3** Performance improvement of ER-BLS on SCUT-FBP5500

| Model | Decreased time(s) | Improved AC (%) |
|---|---|---|
| InceptionResNetV2 | 24,604.99 | 3.00 |
| *EfficientNetB7* | *16,297.85* | *1.99* |
| InceptionV3 | 10,338.87 | 1.53 |
| DensNet121 | 8570.69 | 3.92 |
| ResNet50 | 6919.19 | 3.46 |
| Xception | 5478.81 | 1.71 |
| NASNet | 4201.23 | 6.58 |
| MobileNetV2 | 2932.59 | 5.66 |
| *E-BLS (ours)* | *108.05* | *1.56* |
| BLS | − 651.33 | 8.84 |

**Table 4** Performance improvement of ER-BLS on LSAFBD

| Model | Decreased time(s) | Improved AC (%) |
|---|---|---|
| InceptionResNetV2 | 34,566.99 | 4.82 |
| InceptionV3 | 22,914.75 | 3.71 |
| EfficientNetB7 | 18,879.07 | 4.22 |
| ResNet50 | 17,727.51 | 3.76 |
| DensNet121 | 17,233.74 | 2.20 |
| Xception | 14,566.06 | 5.12 |
| NASNet | 9757.41 | 9.01 |
| MobileNetV2 | 5543.93 | 7.39 |
| *E-BLS (ours)* | *13.94* | *1.31* |
| BLS | − 840.17 | 9.17 |

**Table 5** ER-BLS with different feature nodes and enhancement nodes on SCUT-FBP5500

| Feature nodes | Enhancement nodes | Training time (s) | Testing AC (%) |
|---|---|---|---|
| 12 × 54 | 1000 | 1292.1 | 73.77 |
| 12 × 54 | 2000 | 1294.65 | 74.226 |
| 12 × 54 | 3000 | 1294.71 | 74.59 |
| 12 × 54 | 4000 | 1295.97 | 74.317 |
| 12 × 54 | 5000 | 1295.95 | 74.317 |
| 12 × 54 | 6000 | 1296.36 | 73.678 |
| 12 × 54 | 7000 | 1295.75 | 73.133 |
| 12 × 54 | 8000 | 1297.60 | 72.951 |
| 14 × 54 | 8000 | 1300.92 | 72.222 |
| 16 × 54 | 8000 | 1303.64 | 72.86 |
| 16 × 57 | 8000 | 1306.09 | 72.86 |
| 16 × 60 | 8000 | 1305.21 | 72.678 |

**Table 6** ER-BLS with different number of training samples on LSAFBD

| Training samples | Training time (s) | Testing AC (%) | Training AC (%) |
|---|---|---|---|
| 4000 | 2280.87 | 51.30 | 93.45 |
| 5000 | 2282.06 | 51.40 | 89.10 |
| 6000 | 2283.64 | 55.11 | 81.17 |
| 7000 | 2285.04 | 58.72 | 77.19 |
| 8000 | 2286.54 | 62.13 | 72.34 |

### 4.7 Incremental learning

ER-BLS can complete model construction and training by incremental learning. It is a dynamic system whose performance would be improved by adding additional feature nodes, enhancement nodes or new data. At the same time, the structure of ER-BLS would be quickly updated and trained, which greatly improves the efficiency of ER-BLS. To verify the incremental learning effect of the presented

3. We infer that ER-BLS with transfer learning is superior to that without transfer learning, which suggests that prior knowledge can improve the performance of FBP.

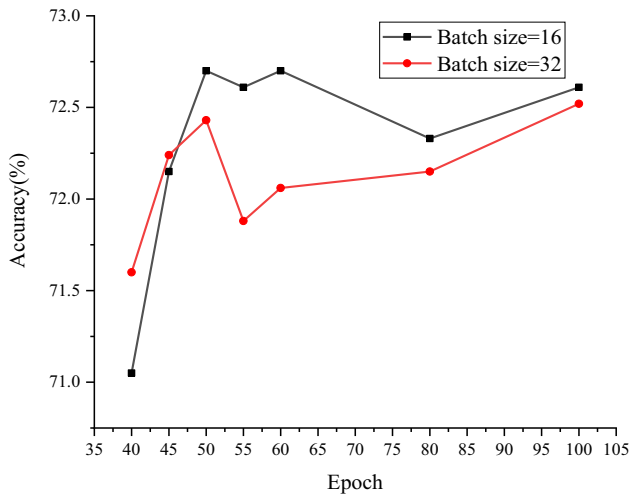

**Fig. 9** The accuracy of EfficientNetB7 with different training epoch and batch size for FBP on SCUT-FBP5500

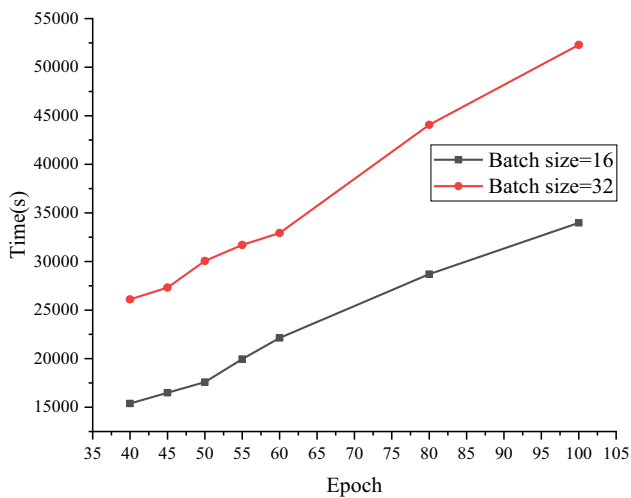

**Fig. 10** The training time of EfficientNetB7 with different training epoch and batch size for FBP on SCUT-FBP5500

methods, we conducted experiments on SCUT-FBP5500 and LSAFBD by adding feature nodes and enhancement nodes. For SCUT-FBP5500, the initial feature nodes are 548 and enhancement nodes are 500. For LSAFBD, ER-BLS was initialized with 1700 feature nodes and 1000 enhancement nodes. Then, the incremental algorithm was adopted to add dynamically 20 feature nodes and 500 enhancement nodes each time. The testing results of incremental learning are listed in Tables 8 and 9. For SCUT-FBP5500, ER-BLS is reconstructed and trained within 12 s. For LSAFBD, ER-BLS is reconstructed and trained within 37 s. Compared with deep convolutional neural networks that consume a lot of time for retraining, ER-BLS greatly improved the efficiency of the model.

### 4.8 Comparison with the other methods

To further verify the effectiveness of E-BLS and ER-BLS, the methods presented were compared with the related methods, respectively. The results are listed in Tables 10 and 11. For SCUT-FBP5500, the Pearson correlation of E-BLS is 0.9104 and ER-BLS is 0.9303, which is the best among various methods. For LSAFBD, testing accuracy of E-BLS is 60.82% and ER-BLS is 62.13%, which is the best among various methods. E-BLS and ER-BLS presented can be extended to the fields of pattern recognition, object detection and image classification.

## 5 Conclusion

FBP is a challenge task because of the complexity of human perception, the subjectivity of human aesthetics, and the diversity of human appearance. In this paper, we present a new novel scheme, BLS fused transfer learning, for FBP task to improve its real-time performance. Meanwhile, we instantiate two fusing networks E-BLS and ER-BLS. And extensive ablation studies confirm the effectiveness and superiority of E-BLS and ER-BLS on two databases. Traditionally, deep networks depend on high-performance PC and take training time of hours. The methods proposed enable the establishment of a high accuracy FBP model in a normal PC within 40 min.

**Table 7** Comparison of ER-BLS with different backbones and Transfer Learning on SCUT-FBP5500

| Backbones | Transfer learning | Training time (s) | Testing AC (%) | PC |
|---|---|---|---|---|
| EfficientNetB1 | No | 791.05 | 61.65 | 0.7680 |
|  | Yes | 689.09 | 71.59 | 0.9085 |
| EfficientNetB3 | No | 1033.91 | 63.66 | 0.7895 |
|  | Yes | 881.34 | 72.13 | 0.9127 |
| EfficientNetB5 | No | 1262.44 | 62.75 | 0.8361 |
|  | Yes | 1031.13 | 73.22 | 0.9209 |
| EfficientNetB7 | No | 1426.64 | 61.75 | 0.8763 |
|  | *Yes* | *1291.83* | *74.69* | *0.9303* |

Table 8 Incremental learning results of ER-BLS on SCUT-FBP5500

| Feature nodes | Enhancement nodes | Training time (s) | Testing accuracy (%) |
| --- | --- | --- | --- |
| 548 | 500 | 5.18 | 73.32 |
| 548–568 | 500–1000 | 8.39 | 73.50 |
| 568–588 | 1000–1500 | 9.49 | 73.68 |
| 588–608 | 1500–2000 | 10.12 | 73.59 |
| 608–628 | 2000–2500 | 11.19 | 74.14 |
| 628–648 | 2500–3000 | 11.98 | 73.86 |

Table 9 Incremental learning results of ER-BLS on LSAFBD

| Feature nodes | Enhancement nodes | Training time (s) | Testing AC (%) |
| --- | --- | --- | --- |
| 1700 | 1000 | 19.54 | 60.62 |
| 1700–1720 | 1000–1500 | 26.52 | 61.12 |
| 1720–1740 | 1500–2000 | 28.49 | 60.87 |
| 1740–1760 | 2000–2500 | 30.99 | 61.17 |
| 1760–1780 | 2500–3000 | 32.94 | 61.32 |
| 1780–1800 | 3000–3500 | 36.45 | 61.42 |

Table 10 Performance comparison with the other methods on SCUT-FBP5500

| Model | PC |
| --- | --- |
| P-AaNet (Lin et al. 2019a, b) | 0.8965 |
| 2 M BeautyNet (Gan et al. 2020a, b) | 0.8996 |
| AestheticNetG (Danner et al. 2021) | 0.9011 |
| AaNet (Lin et al. 2019a, b) | 0.9055 |
| E-BLS (ours) | 0.9104 |
| MSMFME (Dornaika and Moujahid 2022) | 0.9113 |
| R$^3$CNN (Lin et al. 2019a, b) | 0.9142 |
| REX-INCEP (Bougourzi et al. 2022) | 0.9165 |
| ER-BLS (ours) | 0.9303 |

Table 11 Performance comparison with the other methods on LSAFBD

| Model | Testing AC (%) |
| --- | --- |
| Deep cascaded forest (Zhou and Feng 2017) | 54.29 |
| Multi-scale K-means (Gan et al. 2017) | 55.07 |
| NIN (Szegedy et al. 2015) | 58.30 |
| NetA + DAL (Gan et al. 2019) | 59.90 |
| DeepID2 (Zhai et al. 2019) | 60.25 |
| Noise Labels (Gan et al. 2022a, b) | 60.80 |
| E-BLS (ours) | 60.82 |
| Cross Network (Gan et al. 2022a, b) | 61.29 |
| LDCNN (Gan et al. 2020a, b) | 62.00 |
| ER-BLS (ours) | 62.13 |

In the future, in addition to continuing to improve the performance of our networks, we also will consider psychological conclusions about facial beauty and facial attribute-aware such as gender, race and age that influence facial beauty.

**Acknowledgements** This work was supported in part by the National Natural Science Foundation of China under Grant 61771347, in part by the Guangdong Basic and Applied Basic Research Foundation under Grant 2019A1515010716 and in part by the Basic Research and Applied Basic Research Key Project in General Colleges and Universities of Guangdong Province under Grant 2018KZDXM073.

**Funding** The authors have not disclosed any funding.

**Data availability** Enquiries about data availability should be directed to the authors.

## Declarations

**Conflict of interest** All the authors declare that he/she has no conflict of interest.

**Ethical approval** This paper does not contain any studies with human participants or animals performed by any of the authors.

**Publisher's Note** Springer Nature remains neutral with regard to jurisdictional claims in published maps and institutional affiliations.